%% file: neurips_2025.tex
\documentclass{article}

\usepackage[numbers,square]{natbib}

\usepackage[preprint]{neurips_2025}

\usepackage[utf8]{inputenc} 
\usepackage[T1]{fontenc}    
\usepackage{hyperref}       
\usepackage{url}            
\usepackage{booktabs}       
\usepackage{amsfonts}       
\usepackage{nicefrac}       
\usepackage{microtype}      
\usepackage{xcolor}         
\usepackage{amsmath}
\usepackage{wrapfig}
\usepackage[ruled,vlined,linesnumbered,nosemicolon]{algorithm2e}
\usepackage{graphicx}  
\usepackage{adjustbox}
\usepackage{relsize}        
\newcommand*\acr[1]{\textscale{.9}{#1}} 
\usepackage{enumitem}
\usepackage{array}
\usepackage{bm}
\usepackage{booktabs}
\usepackage{graphicx}
\usepackage{graphics}
\usepackage{mathrsfs}
\usepackage{microtype}
\usepackage{multicol}
\usepackage{ragged2e}
\usepackage{relsize}
\usepackage{setspace}
\usepackage{subfigure}
\usepackage{tikz}
\usepackage{pgf}
\usepackage{pgfplots}
\usepackage{xcolor}
\usepackage{pgfplotstable}
\usepackage{siunitx}  
\usepackage{listings}
\usepackage{tcolorbox}

\usepackage{listings}
\usepackage{tcolorbox}

\input{math_commands.tex}

\newtcolorbox[auto counter, number within=section]{tracebox}[1][]{
    coltitle=white,
    colbacktitle=b4!90, 
    colback=gray!10,    
    colframe=black,     
    titlerule=0.5pt,
    fontupper=\ttfamily\tiny, 
    boxsep=7pt,
    boxrule=0.9pt,      
    arc=4pt,            
    width=\linewidth,   
    title={Main Agent Prompt}, 
}

\SetKwProg{Fn}{}{}{end}
\SetKwComment{Comment}{$\triangleright$\ }{}

\definecolor{light-gray}{gray}{0.95}

\pgfplotsset{compat=1.16}
\usepgfplotslibrary{external}
\usepgfplotslibrary{fillbetween}
\tikzstyle{every plot}=[prefix=plots/]

\newcommand\inputpgf[2]{{
\let\pgfimageWithoutPath\pgfimage
\renewcommand{\pgfimage}[2][]{\pgfimageWithoutPath[##1]{#1/##2}}
\let\includegraphicsWithoutPath\includegraphics
\renewcommand{\includegraphics}[2][]{\includegraphicsWithoutPath[##1]{#1/##2}}
\input{#1/#2}
}}

\usetikzlibrary{automata}
\usetikzlibrary{arrows,automata}
\usetikzlibrary{arrows.meta}
\usetikzlibrary{backgrounds}
\usetikzlibrary{cd}
\usetikzlibrary{decorations.markings}
\usetikzlibrary{decorations.pathmorphing}
\usetikzlibrary{intersections}
\usetikzlibrary{positioning}
\usetikzlibrary{math}
\usetikzlibrary{fit,shapes.multipart}

\definecolor{econblue}{HTML}{076FA1} 
\definecolor{econred}{HTML}{ED1F21} 

\definecolor{b1}{HTML}{00526D} 
\definecolor{b2}{HTML}{00A4DC} 
\definecolor{b3}{HTML}{70D0F6} 

\definecolor{b4}{HTML}{0E1D24} 
\definecolor{b5}{HTML}{1B3758} 
\definecolor{b6}{HTML}{3F7CA4} 
\definecolor{b7}{HTML}{53A6E5} 

\definecolor{l1}{HTML}{973d4c} 
\definecolor{l2}{HTML}{AC8B96} 
\definecolor{l3}{HTML}{30c1d3} 
\definecolor{l4}{HTML}{076FA1} 

\definecolor{c1}{HTML}{F2836B} 
\definecolor{c2}{HTML}{F2BFAC} 
\definecolor{c3}{HTML}{6394BF} 
\definecolor{c4}{HTML}{314259} 
\definecolor{c5}{HTML}{3B5E8C} 

\title{A Self-Improving Coding Agent}

\author{%
    Maxime Robeyns, \\
  University of Bristol \\
  \texttt{maxime@igent.ai} \\
  \And
  Martin Szummer \\
  iGent AI \\
  \texttt{szummer@igent.ai} \\
  \AND
  Laurence aitchison \\
  University of Bristol \\
  \texttt{laurence.aitchison@bristol.ac.uk} \\
}

\begin{document}

\maketitle

\begin{abstract}
Recent advancements in Large Language Models (LLMs) have spurred interest in
deploying LLM agents to undertake tasks in the world. LLMs are often deployed in
agent systems: code that orchestrates LLM calls and provides them with tools. We
demonstrate that an agent system, equipped with basic coding tools, can
autonomously edit itself, and thereby improve its performance on benchmark
tasks. We find performance gains from 17\% to 53\% on a random subset of SWE
Bench Verified, with additional performance gains on LiveCodeBench, as well as
synthetically generated agent benchmarks. Our work represents an advancement in
the automated and open-ended design of agentic systems, and demonstrates a
data-efficient, non gradient-based learning mechanism driven by LLM reflection
and code updates.
\end{abstract}

\input{sections/01_introduction}
\input{sections/02_related_work}
\input{sections/03_methods}
\input{sections/04_experiments}

\input{sections/05_conclusion}


{
\small

\bibliographystyle{plainnat}
\bibliography{references}

}


\appendix

\input{sections/appendix}


\end{document}

%% file: math_commands.tex
\usepackage{amsmath,amsfonts,bm}

\def\eqref#1{equation~\ref{#1}}

\def\1{\bm{1}}

\DeclareMathAlphabet{\mathsfit}{\encodingdefault}{\sfdefault}{m}{sl}
\SetMathAlphabet{\mathsfit}{bold}{\encodingdefault}{\sfdefault}{bx}{n}

\DeclareMathOperator*{\argmax}{arg\,max}

%% file: sections/01_introduction.tex
\section{Introduction} \label{sec:introduction}

\acr{LLM}s have recently made impressive
advancements across a range of domains and tasks
\citep{anthropicIntroducingClaude35,googleIntroducingGemini2024,openaiOpenAIO1System}.
However, in order to put these \acr{LLM}s to use in real world applications,
\acr{LLM}s must be wrapped in
code to orchestrate them and expose tools that allow the models to take
actions. These action-taking \acr{LLM}s are referred to as agents, and the
broader system an \emph{agent system}.

These agent systems often show dramatic improvements in benchmark performance
over ``plain'' \acr{LLM}s \citep{yaoReActSynergizingReasoning2023,
zelikmanParselAlgorithmicReasoning2023, chenProgramThoughtsPrompting2023},
through combinations of prompting strategies and methods for combining
different \acr{LLM} outputs. Early examples include best-of-$N$ sampling and
simple prompting strategies such as chain of thought
\citep{kojimaLargeLanguageModels2022}. However more sophisticated schemes have
shown success in getting the desired behavior and
performance improvements from the models, for instance STaR
\citep{zelikmanSTaRBootstrappingReasoning2022}, Tree of Thoughts
\citep{yaoTreeThoughtsDeliberate2023}, Graph of Thoughts
\citep{bestaGraphThoughtsSolving2023}, \acr{LLM} Debate
\citep{duImprovingFactualityReasoning2023}, Iterative Self-Refinement
\citep{madaanSelfRefineIterativeRefinement2023}, Expert Prompting
\citep{longMultiexpertPromptingImproves2024} among many others.
The comprehensive survey of \citet{schulhoffPromptReportSystematic2024}
demonstrates the vast number of manually created strategies to date.

Recent improvements in coding agents raise the question of whether these agents themselves can autonomously modify
and improve their own code by discovering e.g.\ new prompting schemes or tools without manual design and implementation.
We argue that this style of fully self-referential meta-agent programming is
possible today and offers a sound alternative to the ad-hoc, trial-and-error
approach of hand-crafted orchestrators, which may only explore a small fraction
of the solution space.
Recent work in the Automated Design of Agentic Systems (\acr{ADAS})
\citep{huAutomatedDesignAgentic2024} uses a meta-agent to optimise agent
implementations.
However, \citet{huAutomatedDesignAgentic2024} is not \textit{self}-improving, as
there are two separate agents: the target-agent that performs the task, and the
meta-agent, which improves the target agent. A motivation for a self-improving
system is that the improvements in coding abilities may be leveraged during
subsequent improvement steps, hopefully compounding.

Our contributions are:
\vspace{-5pt}
\begin{itemize}
\item  A self-improving coding agent (\acr{SICA}) that eliminates the
distinction between meta-agent and target agent, and is capable of editing its own codebase to
improve itself with respect to its cost, speed and benchmark performance.
\item Empirical evidence that self-referential agents can effectively improve
      their own implementations; we find performance improves from 17\% to 53\% performance on a random subset of SWE-Bench Verified, even with consideration given to safety constraints and
      resource efficiency.
\item We share a our implementation of a self-improving coding agent (\acr{SICA}) with the community.  SICA is implemented in standard Python without a domain-specific language, and provides a reference agent framework for building new SICA systems, as well as those seeking to post-train LLMs on tool use and other agent tasks.
\end{itemize}

We make our code available at {\small \url{https://github.com/MaximeRobeyns/self\_improving\_coding\_agent}}.

%% file: sections/02_related_work.tex
\section{Related Work}%
\label{sec:related_work}

\begin{figure}[t]
    \centering
    \includegraphics[width=\linewidth]{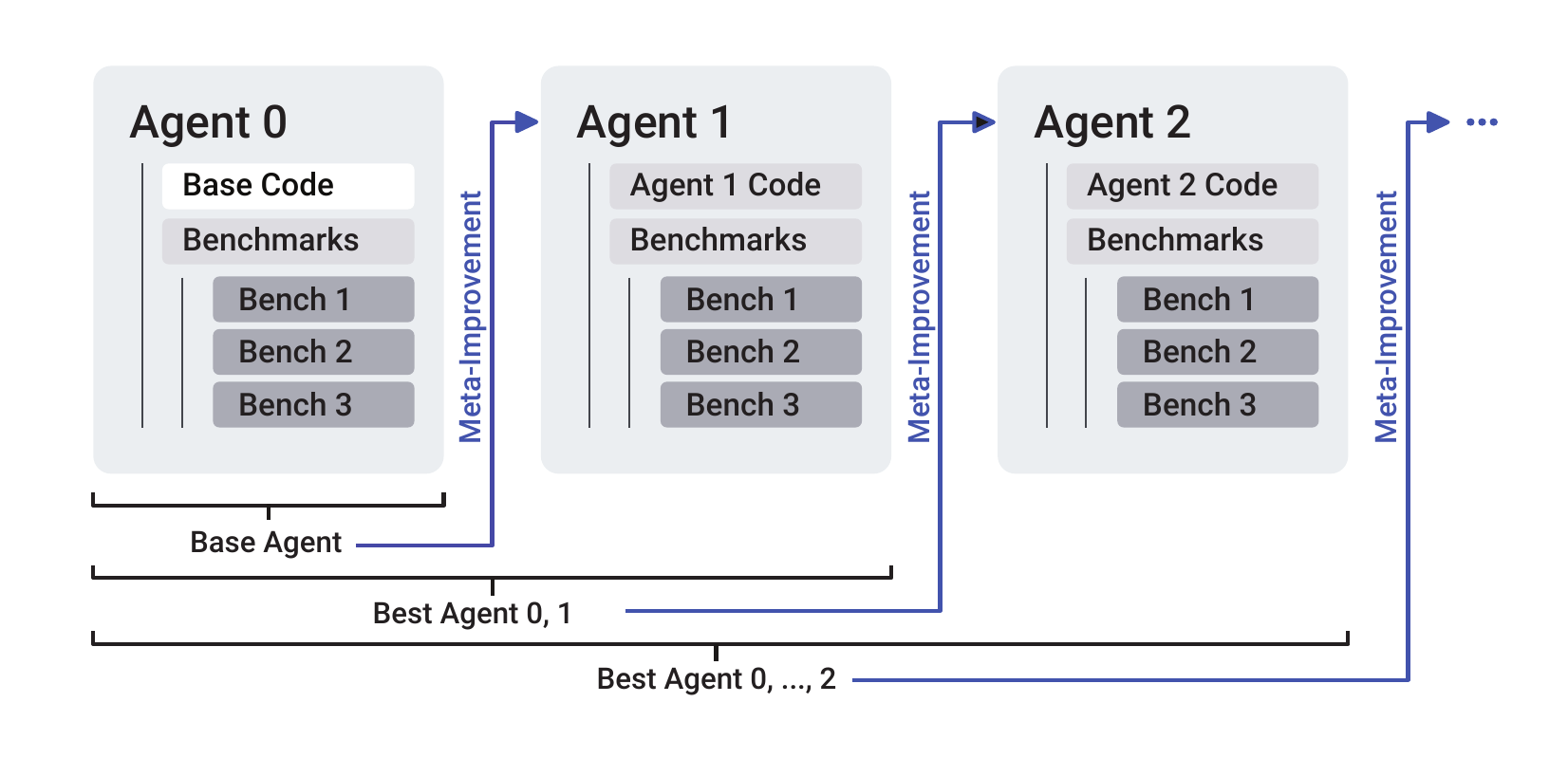}
    \caption{Meta Agent Loop: the agents starts with the minimal code required
      to support initial self-improvement, and then follows a sequence of
      benchmarking and meta-improvement.}
    \label{fig:agent_loop}
\end{figure}

The traditional approach to developing and optimizing agent systems has been to manually design agent architectures and
prompting techniques. Notable examples include Chain-of-Thought prompting
\citep{weiChainofThoughtPromptingElicits2022}, self-refinement
\citep{madaanSelfRefineIterativeRefinement2023} and self-reflection
\citep{shinnReflexionLanguageAgents2023} for improving reasoning, tool use
frameworks \citep{schickToolformerLanguageModels2023}, and various compositional
agent systems \citep{wangVoyagerOpenEndedEmbodied2023, ahnCanNotSay2022}. While these hand-crafted
approaches have achieved strong results, they require significant human effort
and may miss useful patterns that could be discovered through automated search.

Another direction focuses on enabling agents to learn reusable skills and
continuously self-improve. MaestroMotif
\citep{klissarovMaestroMotifSkillDesign2024} uses LLM feedback to
learn skill rewards and combines skills through code generation. This builds on
earlier work on intrinsically motivated reinforcement learning \citep{chentanezIntrinsicallyMotivatedReinforcement2004}
and autotelic agents \citep{colasAutotelicAgentsIntrinsically2022} that develop
repertoires of internally motivated skills, as well as work in open-endedness
\cite{zhangOMNIOpenendednessModels2024, faldorOMNIEPICOpenendednessModels2024}
that use \acr{LLM}s to identify interesting and useful directions to explore in.

Several approaches leverage LLMs to optimize agent behaviors through natural
language interaction. OPRO \citep{yangLargeLanguageModels2024} and Promptbreeder
\citep{fernandoPromptbreederSelfReferentialSelfImprovement2023} focus on
optimizing prompts through language. Others have explored using LLMs to critique
and refine agent behaviors \citep{klissarovMotifIntrinsicMotivation2023},
generate training curricula \citep{kumarPracticeMakesPerfect2024}, or provide
natural language feedback for reinforcement learning
\citep{quRecursiveIntrospectionTeaching2024}.
Since our agent can edit its entire codebase, this includes the ability to tune its own prompts to optimise
the behaviour of any part of the agent.

Recent work has begun exploring automated approaches for designing and
optimizing agent systems. AgentSquare \citep{shangAgentSquareAutomaticLLM2024}
proposed a modular design space that abstracts agent components into planning,
reasoning, tool use and memory modules, allowing automated search over module
combinations.
AlphaEvolve \citep{novikovAlphaEvolveCodingAgent2025} set out to use a coding agent to
make scientific discoveries and optimize computational infrastructure, the
results of which may accelerate the training of the LLM underpinning
AlphaEvolve itself. The approach taken by these works leans more heavily on
structured, evolutionary search than our own.

Perhaps the most closely related line of prior work began with
\acr{ADAS} \citep{huAutomatedDesignAgentic2024}.
\acr{ADAS} used a target-agent which performs the actual task, and a meta-agent which improves the target-agent.
As such, \acr{ADAS} is not self-improving (as the meta-agent improves the target agent, not itself).
Moreover, in \acr{ADAS} the meta-agent edits only a single \texttt{forward} function, written in a domain-specific language which has been carefully designed to make expressing different prompting schemes very straightforward.
In contrast, our self-improving coding agent is fully self-improving (i.e.\
there is no distinction between the meta and target agent), and it operates over
the agent's full Python codebase.

Of course, we would expect the first truly self-improving agents to be coding agents, because agents are written in code.
The natural approach is to start off with a basic coding agent that can open/close/edit files, run commands in the terminal etc, then to launch this agent in a self-improvement loop.
We believe that our self-improving coding agent is the first such work.
However, there are two papers claiming self-improving agents, but they do not evaluate in the coding setting, as they do not consider ``full'' coding agents.
First, Gödel Agent
\citep{yinGodelAgentSelfReferential2024a} has specific tools (such as \verb|action_adjust_logic| and \verb|action_read_logic|) that allow modification of small parts of the agent as it is running.
Thus, it is not a general-purpose coding agent, as it is traditionally understood.
And as such, as with \acr{ADAS}, it was evaluated on language understanding and mathematical benchmarks (DROP \citep{dua2019drop}, MGSM \citep{shi2022language}, MMLU \citep{hendrycks2020measuring} and GPQA \citep{reinGPQAGraduateLevelGoogleProof2023}), rather than coding benchmarks.
Second,
\citet{zelikmanSelfTaughtOptimizerSTOP2024} introduce a self-taught optimizer
for recursively self-improving code generation.
Again, this is not a general-purpose coding agent with the ability to e.g.\ work on arbitrary software engineering tasks, and as such, it self-improves and evaluates and only on algorithmic tasks such as learning parity with noise \citep{blum2003noise},
String Grid Distance, and 3-SAT.

%% file: sections/03_methods.tex
\section{Methods}
\label{sec:methods}

The main running loop of \acr{SICA} resembles the
\acr{ADAS} loop \citet{huAutomatedDesignAgentic2024} and is illustrated in
Figure~\ref{fig:agent_loop}.
In particular, both \acr{SICA}, and \acr{ADAS} keep an archive of previous agents and their benchmark results.
In the first step, \acr{SICA} takes the best performing agent from the archive up until this point as the meta-agent, which is instructed to
look through the archive, identify an improvement, and implement it. The
high-level scheme is illustrated in Algorithm~\ref{algo:adas}.
Note that this is different from \acr{ADAS}, in that \acr{ADAS} has a fixed meta-agent, so does not select the meta-agent from the archive (the archive in \acr{ADAS} is an archive of target-agents, not meta-agents).
Second, both \acr{ADAS} and \acr{SICA} evaluate the new agent on a set of evaluation benchmarks and store the results.

We define the `best' performing agent by a utility
function, which incorporates common desiderata for a good agent; the normalised
benchmark performance score $p_{\text{score}} \in [0, 1]$, the wall-clock time
in seconds $p_{\text{time}}$, as well as the dollar cost $p_{\text{cost}}$. The
base utility is calculated as
\begin{equation}
\label{eq:base_utility}
U = w_{\text{score}} p_{\text{score}} + w_{\text{cost}} (1 - \min(1, p_{\text{cost}} / \$ 10)) + w_{\text{time}} (1 - \min(1, p_{\text{time}} / 300 \text{ s}))
\end{equation}
where we set the coefficients to $w_{\text{score}} = 0.5$,
$w_{\text{cost}} = 0.25$ and $w_{\text{time}} = 0.25$.
We cap the worst cost at
\$10 per problem, and impose a timeout of 300 s after which we cancel the agent.
To give partial credit for work done before the timeout, we calculate the final
utility as follows, with a timeout penalty of $\tau = 0.5$:
\begin{equation}
U_{\text{final}} =
\begin{cases}
    U (1 - \tau), & \text{if timed out} \\
    U, & \text{otherwise}
\end{cases}
\end{equation}

Note that since we are not performing any weight updates, this numerical score
is only used to pick the next meta agent, as well as the base agent for the next
iteration.

We begin by describing the initial coding agent.
We then go on to describe the benchmark runner framework, and how this framework
naturally allows us to create a self-referential task (that of improving the
coding agent).

\begin{wrapfigure}{R}{0.35\textwidth}
\begin{tikzpicture}[
    box/.style={
        draw,
        rectangle,
        minimum width=5cm,    
        align=left,
        rounded corners=2pt,
        fill=white
    },
    section/.style={
        draw,
        rectangle,
        minimum width=4cm,    
        align=left,
        rounded corners=2pt,
        fill=b4!20
    },
    component/.style={
        rectangle,
        align=left,
        text width=4cm       
    }
]

\node[section] (system) at (0,0) {System Prompt};
\node[component, below=0.55cm of system.north] (system_comp) {
    \textbullet~Agent definition and goals\\
    \textbullet~Tool documentation\\
    \textbullet~Sub-agent documentation\\
    \textbullet~System instructions
};

\node[section, below=2cm of system] (core) {Core Prompt};
\node[component, below=0.55cm of core.north] (core_comp) {
    \textbullet~Problem statement or instruction\\
    \textbullet~Open files\\
    \textbullet~Directory tree
};

\node[section, below=1.7cm of core] (assistant) {Assistant Messages};
\node[component, below=0.55cm of assistant.north] (assistant_comp) {
    \textbullet~CoT reasoning\\
    \textbullet~Tool calls and results\\
    \textbullet~Sub-agent calls and results\\
    \textbullet~Overseer notifications
};

\begin{scope}[on background layer]
    \node[box,fit={(system) (system_comp) (core) (core_comp) (assistant) (assistant_comp)},inner sep=0.5cm] (container) {};
\end{scope}

\node[rotate=90,anchor=south] at (container.west) {LLM Context Window};

\end{tikzpicture}
\label{fig:context_window}
\caption{LLM context window structure.}
\end{wrapfigure}
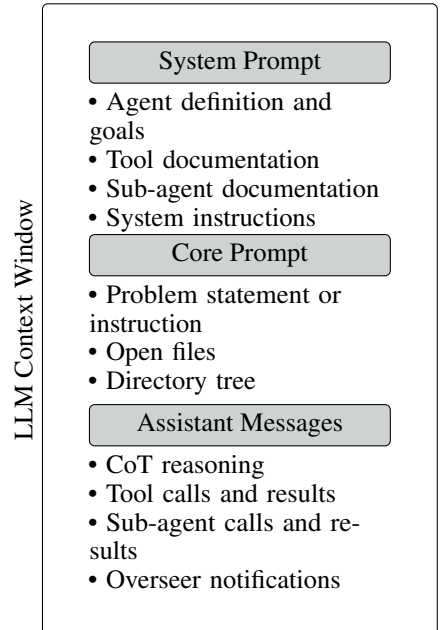

\subsection{Initial agent}%
\label{subsec:tools}

We were interested in useful, practically relevant coding agents capable of self-improvement.
To that end, we started with a sophisticated coding agent, with a number of tools, sub-agents and oversight mechanisms.
Of course, the tools, sub-agents and oversight mechanisms were all amenable to self-improvement; but here we simply describe the starting point.

We designed a set of tools to open (i.e.~add to the context) and close files, overwrite files, execute a shell command and a simple calculator tool. We also include basic tools to submit a final answer to a benchmark problem, return a result from a sub-agent call to its caller, as well as exiting early from a sub-agent call. Finally, we provide an `archive analysis' tool, which facilitates extracting high-level summary statistics about each agent iteration in the archive.
We describe the tool calling mechanism in more detail in Appendix~\ref{sec:function_calling}.

Next, we included several \emph{sub-agents}. These are merely agents invoked by
an existing agent, much like a tool call, by accepting input arguments and
returning a string value.
Introducing this abstraction serves two purposes.
The first is to ensure the benchmark tasks can run entirely autonomously, without human direction and guidance.
We provide the problem statement to the main entrypoint agent, (the full prompt for which is given in Appendix~\ref{subsec:prompts})
which then decomposes the task, delegating and asking questions to sub-agents.
The second is to manage the context length for extended agent runs. The meta-improvement task is challenging, and the main agent may take on the order of one hour to finish one meta-improvement iteration. The sub-agent calls allow the context used for a sub-task to be naturally discarded when the agent returns, with the value returned by the sub-agent providing a summary of that sub-task in the main agent's context.
Initially, we included three agents: a coding agent, a problem-solver agent and a reasoning agent.
The full prompts for these agents are given in Appendix~\ref{subsec:sub_agent_prompts}.


Finally, one important component we provide in the initial agent that enables
extended, autonomous operation while remaining on-task, avoiding loops or other
behaviours that frustrate progress is the asynchronous overseer. This is an
\acr{LLM}, running periodically in a concurrent thread, which is prompted to
judge whether there are any pathological behaviours or deviations from the
original task, and which may intervene by notifying the currently running agent
(or any of its parents) by means of a message in-lined into the assistant
context, or in serious cases to cancel the execution of an agent if it has
deviated too far or appears stuck. The overseer is called every 30s by default,
or on key events.

This overseer is provided with a comprehensive textual representation of the
state of the system (a truncated example of which is provided in
Appendix~\ref{subsec:traces}) with verbose content truncated to limit context.
This consists of the `callgraph' of agent and sub-agent execution, in addition
to the event stream of \acr{LLM} messages, tool calls and responses, and
sub-agent calls and responses associated with each agent call.  This broad
representation of the system's behaviour, allows the overseer to to pick out
patterns, as well as detect repeated work completed by previous agent calls. The
full Overseer prompt is given in Appendix~\ref{subsec:overseer_prompt}.

The structure of the agent's contexts, which contains e.g.\ open file contents, rather than just the prompt is critical.  In the initial coding agent, the context is structured as Figure~\ref{fig:context_window}.
The system prompt with the agent definition (from Appendix~\ref{subsec:prompts}) is presented first, listing the
definitions of the tools that the agent has available and
the sub-agents it has available to call too. The system prompt concludes with system
information, such as instructions for how to break out of the agent loop and
return to the calling process.

Next comes the `core prompt', which is set as the first user message in the
chat template format\footnote{We use the user message role in the chat template
for the core prompt not by design, but owing to the fact that we are using chat
endpoints from model provider APIs. If fine-tuning the \acr{LLM} specifically
for agentic applications, one might choose a more appropriately named section
delimiter such as \texttt{<core>}.}, and contains the caller-specified problem
statement to work on (the caller might be the user invoking the agent, or an
agent calling a sub-agent). This is also where we insert a view of the files the
agent has open, as well as a representation of the current working directory.

Finally, the remainder of the \acr{LLM} agent's context is the running assistant
message. This contains periods of chain-of-thought reasoning about the problem,
punctuated by tool calls and their responses, sub-agent calls and their
responses, as well as any notifications or comments from the asynchronous
overseer system.

The \acr{LLM} context is laid out in this way to allow new content to be
appended, hence preserving the \acr{KV} cache to the benefit of latency between
tool calls and cost. For the base agent, we adopt a simple scheme where file
edits are appended as diffs, and these are periodically re-consolidated into the
stale file view after a certain number accumulate. More sophisticated \acr{KV}
cache preservation schemes may be written by the agent to drive down cost.

\begin{algorithm}[t]
  \KwIn{Evaluation benchmarks $B = \{b_{j} : j = 1, \ldots, m\} $, iteration count $n \in \mathbb{N}$}
  \KwOut{Improved agent system $A_{n}$}
  Initialise agent $A_{0}$\;
  \For {$i = 0, 1, \ldots, n-1$} {
    Evaluate $A_{i}$ on benchmarks $B$ and store in $p_{i}$\;
    Run $A_{\hat{i}}$ to generate $A_{i+1}$, using the archive of $A_{\le i}$
    and $p_{\le i}$, where $\hat{i} = \argmax_{j \in \{0, \ldots, i\}}p_{j}$\;
  }
  \Return $A_{n}$\;
  \caption{\label{algo:adas}Self-Referential Agent Improvement.}
\end{algorithm}

%% file: sections/04_experiments.tex
\section{Experiments and Results}%
\label{sec:experiments}

\pgfplotsset{
    performance/.style={
        width=\linewidth,
        height=0.6\linewidth,
        grid=major,
        grid style={line width=0.2pt, draw=gray!15},
        axis lines=left,
        xlabel=Agent Iteration,
        ylabel=Utility,
        xlabel near ticks,
        ylabel near ticks,
        legend pos=south east,
        legend cell align=left,
        legend style={
            draw=none,
            fill=white,
            fill opacity=0.8,
            text opacity=1,
            font=\small
        },
        try min ticks=5,
        cycle list={
            {blue!40, opacity=0.4},
            {red!80!black, thick},
            {blue!20},
        },
        tick style={color=gray!70},
        major grid style={line width=0.2pt,draw=gray!15},
        minor grid style={line width=0.1pt,draw=gray!8},
        every axis label/.append style={font=\small},
        tick label style={font=\small}
    }
}

\begin{figure}[t]
    \centering
    \begin{tikzpicture}
    \begin{axis}[performance]
        \pgfplotstableread{data/main_plot_data.dat}\perfdata

        \addplot[c3!90] table[x=iteration,y=target_score] {\perfdata};
        \addlegendentry{Mean Utility}

        \addplot table[x=iteration,y=cummax_score] {\perfdata};
        \addlegendentry{Best Utility So Far}

        \addplot[name path=upper,draw=none] table[x=iteration,y=ci_upper] {\perfdata};
        \addplot[name path=lower,draw=none] table[x=iteration,y=ci_lower] {\perfdata};
        \addplot[c3!30] fill between[of=upper and lower];
        \addlegendentry{95\% Confidence Interval}

        \node[anchor=south, fill=white, fill opacity=0.9, text opacity=1,
              rounded corners=2pt, inner sep=2pt, font=\footnotesize, align=center]
              at (axis cs:1.1,0.63) {`Smart Edit'\\Tool};
        \draw[->,>=stealth,thick,opacity=0.6] (axis cs:1.0,0.63) -- (axis cs:1,0.59);

        \node[anchor=center, fill=white, fill opacity=0.9, text opacity=1,
              rounded corners=2pt, inner sep=2pt, font=\footnotesize, align=center]
              at (axis cs:5.0,0.55) {Code Context\\Summarization};
        \draw[->,>=stealth,thick,opacity=0.6] (axis cs:5.0,0.562) -- (axis cs:5,0.59);

        \node[anchor=center, fill=white, fill opacity=0.9, text opacity=1,
              rounded corners=2pt, inner sep=2pt, font=\footnotesize, align=center]
              at (axis cs:7,0.655) {File Edit\\Verification};
        \draw[->,>=stealth,thick,opacity=0.6] (axis cs:7,0.645) -- (axis cs:7,0.605);

        \node[anchor=center, fill=white, fill opacity=0.9, text opacity=1,
              rounded corners=2pt, inner sep=2pt, font=\footnotesize, align=center]
              at (axis cs:9,0.685) {AST Symbol\\Locator};
        \draw[->,>=stealth,thick,opacity=0.6] (axis cs:9,0.675) -- (axis cs:9,0.65);

        \node[anchor=center, fill=white, fill opacity=0.9, text opacity=1,
              rounded corners=2pt, inner sep=2pt, font=\footnotesize]
              at (axis cs:13,0.7) {Hybrid Symbol Locator};
        \draw[->,>=stealth,thick,opacity=0.6] (axis cs:13,0.693) -- (axis cs:13.9,0.675);
    \end{axis}
    \end{tikzpicture}
    \caption{Performance across iterations. Key improvements are annotated with
      their corresponding tool or agent modifications.}
    \label{fig:performance}
\end{figure}
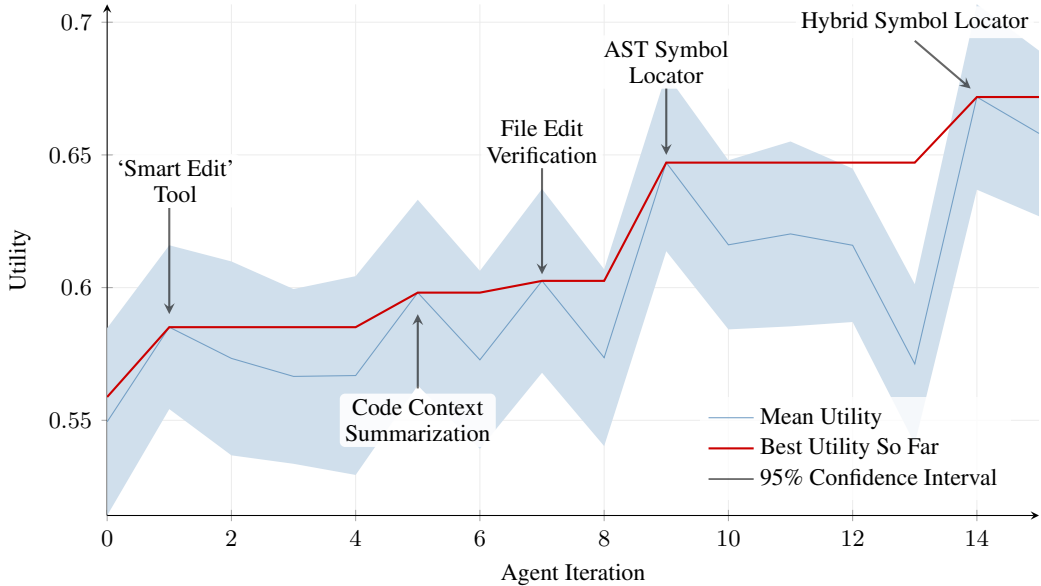

We first show a run that includes relatively long-horizon tasks in the benchmark
set. We include \acr{SWE} Bench Verified
\citep{chowdhuryIntroducingSWEbenchVerified2024} which tests problem
decomposition, code navigation as well as fast and efficient file editing from
which we select a fixed random subset of 50 questions. We also include 50 random
questions from LiveCodeBench~\citep{jain2024livecodebench}, the questions from
which resemble competitive programming and often include more theoretical
reasoning\footnote{Both SWEBench-Verified and LiveCodeBench are MIT licensed}.

We also include two example `synthetic' benchmark tasks, defined and curated
entirely within the agent's codebase. The first of these is a file editing
benchmark, constructed by cloning the repositories used in \acr{SWE} bench,
finding `interesting' (i.e.~not trivial, nor complete overwrites) file edits in
consecutive commits, mounting the repository checked out at the first commit in
the working directory, and prompting the agent to edit the file such that it
matches the target content of the second commit. The agent is scored on the
closeness of the final file content to the target content, with time and dollar
cost and token consumption also being accounted for in the utility function.
The second is a codebase navigation problem, which is also constructed by
cloning some Python repositories, from which we identify `interesting' symbols,
and at least one reference to this symbol in the project. The benchmark task is
for the agent to locate the definition of this symbol, and return the location
in the format \texttt{path/to/file.py:line\_num:column\_num} from the example
reference.

We run the agent system using Sonnet 3.5 (v2)
\citep{anthropicIntroducingClaude35} in most of the agents in the agent system,
with the exception of a `reasoning agent' which uses o3 mini
\citep{openaiOpenAIO3mini2025}. While the base agent system can run with much
smaller and less expensive models, and many of our benchmark tasks can be solved
by this class of models, the
meta-improvement task is a complicated long-horizon task which requires powerful
models. The API cost of the 15 iteration run was approximately \$7,000.

The results are shown in Figure~\ref{fig:performance}, with more detailed
iteration-level peformance numbers in Table~\ref{tab:performance}.

\begin{table}[t]
\caption{Performance per iteration: the benchmark accuracy numbers correspond to
a synthetic file editing benchmark and symbol location (codebase navigation)
benchmark, SWE-Bench Verified, and LiveCodeBench, respectively. We also show the
metrics averaged across all problem instances across all benchmarks to compare
the relative `per problem' costs across agent iterations.}
\label{tab:performance}
\begin{center}

\pgfplotstableread[col sep=space]{data/main_table.dat}\loadeddata

\pgfplotstabletypeset[
    columns={iteration,file,symbol,swe,lcb,cost,time,tokens,cached},
    every head row/.style={
        before row={
            \toprule
            & \multicolumn{4}{c}{\bf Benchmark Accuracy} &
            \multicolumn{4}{c}{\textbf{Average Metrics} (per benchmark problem)} \\
            \cmidrule(lr){2-5} \cmidrule(lr){6-9}
        },
        after row=\midrule
    },
    every last row/.style={after row=\bottomrule},
    columns/iteration/.style={
        column type=r,
        string type,
        column name={\bf Iter}
    },
    columns/file/.style={
        column type=r,
        fixed zerofill,
        precision=2,
        column name={\bf File}
    },
    columns/symbol/.style={
        column type=r,
        fixed zerofill,
        precision=2,
        column name={\bf Sym}
    },
    columns/swe/.style={
        column type=r,
        fixed zerofill,
        precision=2,
        column name={\bf SWE-Bv}
    },
    columns/lcb/.style={
        column type=r,
        fixed zerofill,
        precision=2,
        column name={\bf LCB}
    },
    columns/cost/.style={
        column type=r,
        fixed zerofill,
        precision=2,
        column name={\bf Cost (\$)}
    },
    columns/time/.style={
        column type=r,
        fixed zerofill,
        precision=1,
        column name={\bf Time (s)}
    },
    columns/tokens/.style={
        column type=r,
        string type,
        column name={\bf Tokens (M)}
    },
    columns/cached/.style={
        column type=l,
        string type,
        column name={\bf \% Cached}
    }
]\loadeddata

\end{center}
\end{table}

The numbers show a promising progression. For the file editing benchmark, the
agent is initially only equipped with the ability to overwrite files which is
slow and token inefficient, and it quickly implemented self-termed `smart'
editing tools based on diffs and ranges. For the symbol navigation task, the
agent implemented a simple \acr{AST} based symbol locator at iteration 9
(perhaps exploiting the fact that our benchmarks were predominantly Python
based), which yielded good improvements also reflected in other
tasks.\footnote{Note that we attribute the low score in the symbol locator
  benchmark to poor data quality---some of the target symbols point to
  uninstalled external libraries for instance.} The \acr{SWE} Bench Verified
subset saw an appreciable increase in performance throughout the run, accuring
agent framework improvements. The LiveCodeBench scores saw a subtle
improvement, although not as pronounced as some of the other benchmarks.

\subsection{Performance on Reasoning Tasks}%
\label{subsec:reasoning_tasks}

We also evaluate the effectiveness of the self-referential agent system in
improving task performance in more reasoning-heavy domains. We ran another
experiment with two question answering tasks in the benchmark set: \acr{AIME}
2024 and \acr{GPQA} Diamond \cite{reinGPQAGraduateLevelGoogleProof2023}, the
results of which are shown in Figure~\ref{fig:aime_gpqa}.

\pgfplotsset{
    aime/.style={
        width=\linewidth,
        height=0.4\linewidth,
        grid=major,
        grid style={line width=0.2pt, draw=gray!15},
        axis lines=left,
        title=\acr{AIME} and \acr{GPQA} Show Little Improvement,
        xlabel=Agent Iteration,
        ylabel=Mean Accuracy,
        xlabel near ticks,
        ylabel near ticks,
        legend pos=north east,
        legend cell align=left,
        legend style={
            draw=none,
            fill=white,
            fill opacity=0.8,
            text opacity=1,
            font=\small
        },
        try min ticks=5,
        cycle list={
            {blue!40, opacity=0.4},
            {red!80!black, thick},
            {blue!20},
        },
        tick style={color=gray!70},
        major grid style={line width=0.2pt,draw=gray!25},
        minor grid style={line width=0.1pt,draw=gray!18},
        every axis label/.append style={font=\small},
        tick label style={font=\small}
    }
}

\begin{figure}[htbp]
    \centering
    \begin{tikzpicture}
    \begin{axis}[aime]
        \pgfplotstableread{data/aime_gpqa_plot.dat}\perfdata

        \addplot[c3!90] table[x=iteration,y=target_score] {\perfdata};
        \addlegendentry{Mean Accuracy}

        \addplot table[x=iteration,y=cummax_score] {\perfdata};
        \addlegendentry{Best Mean Accuracy So Far}

        \addplot[name path=upper,draw=none] table[x=iteration,y=ci_upper] {\perfdata};
        \addplot[name path=lower,draw=none] table[x=iteration,y=ci_lower] {\perfdata};
        \addplot[c3!30] fill between[of=upper and lower];
        \addlegendentry{95\% Confidence Interval}

        \node[anchor=south, fill=white, fill opacity=0.9, text opacity=1,
              rounded corners=2pt, inner sep=2pt, font=\footnotesize, align=center]
              at (axis cs:1,0.79) {Independent math\\verifier};
        \draw[->,>=stealth,thick,opacity=0.6] (axis cs:1,0.785) -- (axis cs:1,0.76);

        \node[anchor=north, fill=white, fill opacity=0.9, text opacity=1,
              rounded corners=2pt, inner sep=2pt, font=\footnotesize, align=center]
              at (axis cs:2,0.71) {Sympy symbolic\\calculator};
        \draw[->,>=stealth,thick,opacity=0.6] (axis cs:2,0.715) -- (axis cs:2,0.76);

        \node[anchor=north, fill=white, fill opacity=0.9, text opacity=1,
              rounded corners=2pt, inner sep=2pt, font=\footnotesize, align=center]
              at (axis cs:4,0.66) {Math cross\\validator};
        \draw[->,>=stealth,thick,opacity=0.6] (axis cs:4,0.66) -- (axis cs:4,0.685);

        \node[anchor=south, fill=white, fill opacity=0.9, text opacity=1,
              rounded corners=2pt, inner sep=2pt, font=\footnotesize, align=center]
              at (axis cs:4.8,0.79) {Geometry\\specialist};
        \draw[->,>=stealth,thick,opacity=0.6] (axis cs:5,0.785) -- (axis cs:5,0.77);

        \node[anchor=north, fill=white, fill opacity=0.9, text opacity=1,
              rounded corners=2pt, inner sep=2pt, font=\footnotesize, align=center]
              at (axis cs:6,0.66) {Systematic mathematical\\reasoning};
        \draw[->,>=stealth,thick,opacity=0.6] (axis cs:6,0.66) -- (axis cs:6,0.69);
    \end{axis}
    \end{tikzpicture}
    \caption{Agent Framework Saturation: the benefits the agent system was able
      to find when the models alone (e.g. o3-mini-high) already perform well was
      marginal.}
    \label{fig:aime_gpqa}
\end{figure}
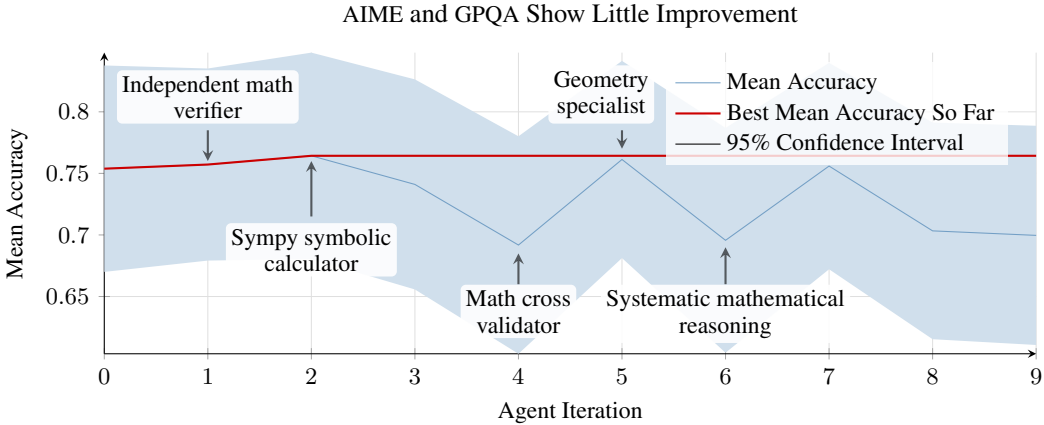

The results here show less improvement, and highlight an important interplay
between the base models and the scaffolding system. The base agent system at
iteration 0 used in Figure~\ref{fig:aime_gpqa} used Sonnet 3.5, with a
`reasoning' sub-agent that used o3-mini. The o3-mini model alone scores 87\% and
79\% on \acr{AIME} and \acr{GPQA} Diamond with a `high' reasoning effort, while
the agent system as a whole averaged 76\% across the two benchmarks.

Inspecting the traces, we observe that for many of the runs, the main agent
merely delegated the problem to the o3-mini based reasoning agent, and did not
leverage the full capabilites of the agent system. For `reasoning models' such
as o1 \citep{openaiOpenAIO1System}, o3-mini or DeepSeek's R1
\citep{deepseek-aiDeepSeekR1IncentivizingReasoning2025}), we suspect that the
inclusion of crude components aiming to induce reasoning behaviour (such as
those included in iterations 4 and 6 in Figure~\ref{fig:aime_gpqa}) may in fact
interrupt the reasoning chain of thought of a reasoning model trained outside
the agent system, resulting in a drop in performance. We look forward to future
work training `agent' \acr{LLM}s jointly with the design of the agent system
which will hopefully make better use of both components' strengths.

%% file: sections/05_conclusion.tex
\section{Conclusion}
\label{sec:conclusion}

We provide SICA, a self-improving coding agent.
Our results demonstrate that self-improvement, implemented via an agent system
editing its own codebase, is indeed possible, particularly in longer-horizon
`agentic' tasks where the base \acr{LLM}s stand to gain more from the structure
and guidance of a scaffolding system---for instance showing improvements from
17\% to 53\% on a random subset of SWE-Bench Verified, while also achieving a
slight decrease in the average time spent per problem on average.
The framework we present offers both a practical approach to automated
agent system design and a foundation for future work in jointly training
foundation models with their agent systems.
\subsection{Limitations}
\label{sec:limitations}

Our initial attempt at a self-improving coding agent is not without
limitations. One key difficulty was getting the LLM-based agent to autonomously
come up with truely novel, innovative, feasible and interesting modification
ideas at each meta-improvement step, which is a theme which has been commented
on in the open-ended learning literature
\citep{nathaniMLGymNewFramework2025,yamadaAIScientistv2WorkshopLevel2025,hughesOpenEndednessEssentialArtificial2024}.
The cost of settling on a bad idea which suffered from poor `taste' was a
lengthy agent editing step followed by an even more expensive run through the
benchmarks. While the failed iteration persists in the archive, in principle
acting as an example of what not to do, we found that the initial feature ideas
would often heavily influence later feature ideas as variations on the same
theme. This path dependency may lead to higher variance agent runs; with poor
quality initial feature suggestions (e.g. fixating on caching open files)
often lowering the quality of subsequent feature suggestions.

We also note that in optimizing for agent running time and cost, our relatively
short 5-minute timeouts (and to a lesser degree per-problem cost limits) cause
the initial benchmark scores to perhaps be lower than expected for the
underlying language model (e.g. Sonnet 3.5 v2), especially for longer-horizon
benchmark tasks like SWE-Bench. Indeed, much of the initial improvement we report
arises from features speeding up file editing and agent cost, which are not
traditionally reflected to the same degree in SWE-Bench numbers.

\subsection{Future Work}
\label{sec:future}

Our approach reflects a non weight-based learning mechanism, where we update the
system's behaviour through a combination of optimising the prompts on which the
model is conditioned, as well as the way the \acr{LLM} calls are orchestrated by
editing the underlying code. This is in some sense very sample-efficient, where
a detailed inspection of the archive of outcomes informs substantial
modifications at each step. However, we recognize the potential of jointly
updating the foundation model's weights with the agent system. The work of
AlphaEvolve \citep{novikovAlphaEvolveCodingAgent2025} demonstrates a proof of
this principle, and it may be fruitful to more explicitly fine-tune the foundation
model to work with new newly created tools and scaffold at each iteration.

While we did not get to the point of benchmark saturation in our relatively
short runs, we expect that future systems may run out of things to learn from a
static benchmark set. One way this problem may be solved is through the
automated design and curation of benchmarks, as initially demonstrated in the
work of \cite{faldorOMNIEPICOpenendednessModels2024} where RL environments are
designed in code. By including two synthetic benchmarks in our work which were
curated entirely by the agent through manual prompting (the file editing and
symbol navigation benchmark) we demonstrate that \acr{SICA} is flexible enough
to support creating new benchmarks and evaluations for itself.

\section{Safety Considerations}%
\label{subsec:safety}

It is vitally important to carefully consider the safety implications posed by systems designed to be self-improving.
While our self-improving coding agent shows improvement in certain agentic tasks, there seem to be limits to what
can be achieved with such purely scaffolding based systems (as seen in Figure~\ref{fig:aime_gpqa}), which do not modify the model weights.
If the system updated the model weights as well as the scaffolding, we would expect that the risks of recursively self-improving systems would be higher. We highlight two key mechanisms for mitigating any risks.

First, a key safety mitigation in such systems is observability: that humans can and do oversee all aspects of potentially self-improving systems.
We have thus architected \acr{SICA} with sophisticated yet user-friendly observability mechanisms, which provide rich insight into the chain-of-thought, actions and sub-agent invocations at each step. These include an interactive web interface for easy oversight of all events occurring in the agent system, as well as the asynchronous LLM-based overseer, which may be prompted to treat various behaviors as grounds for canceling an agent, and which may also use a different LLM to the running agent.
See Appendix~\ref{subsec:overseer_prompt} for an example overseer judgement and
the control afforded to it at each asynchronous invocation.
Given the lack of model weight updates, which make it difficult but not impossible for the model to encode e.g.\ stenographic messages \citep{motwani2023perfect,motwani2024secret} in the chain-of-thought, we believe this provides strong safety mitigation.
That said, for recursively self-improving systems where the model weights are also updated, observability mechanisms, on their own, are clearly inadequate, and further research on safety mechanisms is clearly necessary.

The second opportunity for safety mitigation is inherent to the iterative, evaluation-based update procedure. While our initial benchmark set focuses on measuring capabilities we wish to improve, it is entirely possible to include safety-related evaluations in this benchmark set, to validate each agent before it progresses to the next iteration as the meta-agent.

We do not believe there to be any significant, direct adverse societal
consequences to this work. Our objectives either focus on improving the
mechanics of code editing or the effectiveness of multi-step reasoning through
longer-horizon coding tasks.

%% file: sections/appendix.tex
\section{Agent Prompts}%
\label{subsec:prompts}

The main agent is the entrypoint that routes requests to other agents and
synthesises sub-agent results into the final answer.

{
\begin{adjustbox}{max width=\linewidth, center}
\begin{tracebox}
Your task is to orchestrate sub-agents in order to solve the problem. Here is the problem to solve:

Problem Statement ==============================================================

\{problem\_statement\}

End Problem Statement ==========================================================

You must now delegate this problem to one, two or however many agents are required to thoroughly solve the problem.

Ensure you relay the problem statement accurately and completely to the agent. Sub-agents will have access to the problem statement too, and so it should rarely be necessary to write it out verbatim. In particular, if the problem statement is very long, you should explicitly NOT write it out again in full. Instead, just give clear direction, and trust that the sub-agent can also refer to the problem statement.

You have been given access to file and directory viewing tools, since these can help you get your bearings and direct sub-agents more effectively. These are however meant to help you understand the context in which you operate. You are intentionally not directly equipped with any tools to conduct substantive work, because you are just the router, delegator and orchestrator. The tools that your sub-agents' have available are however listed, and you should carefully refer to these when considering which agent to call next. It is these sub-agents' job to make the necessary state changes to make progress on the task at hand.
\end{tracebox}
\end{adjustbox}
}

\subsection{Base Sub-Agent Prompts}%
\label{subsec:sub_agent_prompts}

{
\begin{adjustbox}{max width=\linewidth, center}
\begin{tracebox}
As a professional and experienced programmer, your approach is to:\\

1. slow down and don't write files fully end-to-end in one go\\
2. first understand your context thoroughly:\\
    2.1 explore the project to locate all the files that could be useful documentation (README.md files, common likely MD documentation files, etc)\\
    2.2 view each of these files, making notes or summaries, and closing irrelevant or long files\\
    2.3 explore the codebase as it relates to your instructions: find all relevant files, in order to identify existing design patterns and conventions\\
3. (optional) prototype and design before starting coding\\
    3.1 come up with some simple toy examples in a testing directory\\
    3.2 use execution feedback to benchmark or compare the approaches\\
    3.3 synthesise the information and learnings into a final design or solution\\
4. make code edits\\
    4.1 identify the most minimal and effective ways to make your required changes\\
    4.2 observe any existing stylistic conventions\\
5. test end-to-end\\
    5.1 Prefer end-to-end testing in testing scripts without test frameworks\\
    5.2 If this is not an option or the project already uses a testing framework, then use that\\
    5.3 Ensure your code is valid, hasn't introduced any regressions and works as intended\\
6. Clean up after yourself\\
    6.1 Check that all the documentation is still up to date after your changes\\
    6.2 Clean up any temporary files or changes\\
\\
NOTE:\\
    - don't create virtual environments\\
    - avoid pytest and mocks if you can; prefer end-to-end scripts\\
    - if the request is clearly exploratory in nature, then you may bypass the rigorous procedure above, and address it appropriately\\
    - call your reasoning agent if you are stuck on a tricky algorithmic or mathematical problem, to help you gain insight and make progress\\
\end{tracebox}
\end{adjustbox}
}

{
\begin{adjustbox}{max width=\linewidth, center}
\begin{tracebox}
Here is the problem you have been asked to solve:\\
\\
Your problem ================\\
\{self.problem\_statement\}\\
==================================\\
\\
Approach:\\
1. Talk over what the question or problem is asking of you\\
2. Plan minimum viable solution and validate approach\\
3. Execute swiftly and adapt based on feedback\\
4. Verify against original requirements\\
\\
Key principles:\\
- Choose simplest viable solution\\
- Validate early and often\\
- Adapt swiftly when blocked\\
\end{tracebox}
\end{adjustbox}
}

{
\begin{adjustbox}{max width=\linewidth, center}
\begin{tracebox}
Here is the initial request, provided for context only, that we are ultimately trying to solve:\\
Initial request:\\
\\
\{self.\_context.initial\_request\}\\
\\
End initial request.\\
\\
Here is the problem you have been asked to solve:\\
\\
Problem statement:\\
\\
\{self.problem\_to\_solve\}\\
\\
End problem statement.
\end{tracebox}
\end{adjustbox}
}

\subsection{Overseer Prompt}%
\label{subsec:overseer_prompt}

{
\begin{adjustbox}{max width=\linewidth, center}
\begin{tracebox}
You are an expert system overseer responsible for monitoring agent execution in real-time. You will be shown execution traces as they develop and must make decisions about whether intervention is needed. You are running in a loop.\\

Your task is to:\\
1. Analyze the current execution state\\
2. Determine if you have enough information to make confident assessments\\
3. Evaluate whether agents are progressing healthily or need intervention\\
4. Provide thoughtful and constructive steering notifications to running agents if intervention is needed\\
5. In extreme cases, cancel the execution of an agent and notify its parent of why this is\\
6. Leave notes for future evaluation iterations if needed\\
\\
You must provide your analysis in the following structured format:\\
\\
{OVERSEER\_EXAMPLES}\\
\\
Here is the current execution trace:\\
\\
{graph\_repr}\\
\\
Previous overseer notes (if any)\\
{self.\_previous\_notes if self.\_previous\_notes else 'No notes.'}\\
\\
Key considerations:\\
    - Agents that have already returned cannot receive notifications\\
    - Look at timestamps and event counts to understand execution progress\\
    - Some agents may legitimately take long periods to complete\\
    - Consider both local agent health and global execution progress\\
    - If the agent appears stuck without terminating, remind it that it can exit by generating <COMPLETE>\{COMPLETE\_STOP\_TOKEN\}\\
    - If, when the agent has been tasked to work on its own code at meta-improvement time, it is attempting to test tools or other features directly (using its own running code), intervene immediately and tell it to either test a new agent end-to-end (python -m workdir.agent -p "<some prompt>"...) or to create test scripts to test the tools directly.\\
    - You may cancel an agent if it is no longer making progress, wasting resources, or very clearly going down the wrong path or deviating from its instructions. You must make at least one attempt to notify the agent of its wrongdoings before issuing a cancellation command.\\
    - When canceling an agent, you must also notify its parent calling agent of the reasons for the cancellation.\\
    - If the task becomes untenable, after multiple failed attempts, and the agent is appearing to repeat itself, instruct the main agent to use its early\_exit tool to exit early.\\
    - Use cancellations sparingly, as they are  disruptive and ultimately more costly than guiding the agent to successful completion, since any new sub-agent will often have to start the task from scratch.\\
\\
The last time you checked was at: {self.\_last\_check\_time.strftime("\%H:\%M:\%S") if self.\_last\_check\_time else "N/A"}\\
The current timestamp is {datetime.now().strftime("\%H:\%M:\%S")}\\
\\
Your analysis:
\end{tracebox}
\end{adjustbox}
}

The judgements of the overseer will generally be in the following form:

{
\begin{adjustbox}{max width=\linewidth, center}
\begin{tracebox}
<OVERSEER\_JUDGEMENT>\\
<making\_progress>false</making\_progress>\\
<is\_looping>true</is\_looping>\\
<currently\_running\_agent>agent\_d4e5f6g7</currently\_running\_agent>\\
<needs\_notification\_reasoning>The agent has been notified previously about its looping behavior but has not corrected course. It's time to forcefully cancel this execution.</needs\_notification\_reasoning>\\
<needs\_notification>true</needs\_notification>\\
<agent\_to\_notify>agent\_b8a796h5</agent\_to\_notify>\\
<notification\_content>Your software developer agent was looping while trying to install the program, and as a result I have cancelled it. You might want to start another software developer with clearer instructions about how to install the program.</notification\_content>\\
<force\_cancel\_agent>true</force\_cancel\_agent>\\
<force\_cancel\_agent\_id>agent\_d4e5f6g7</force\_cancel\_agent\_id>\\
<next\_check\_type>time</next\_check\_type>\\
<next\_check\_delay>10</next\_check\_delay>\\
<notes\_for\_next\_iteration>Forced cancellation of agent\_d4e5f6g7 due to unresponsive looping. Monitor parent agent for proper handling of cancelled child.</notes\_for\_next\_iteration>\\
<OVERSEER\_JUDGEMENT>
\end{tracebox}
\end{adjustbox}
}

\section{Example Traces}%
\label{subsec:traces}

{
\begin{adjustbox}{max width=\linewidth, center}
\begin{tracebox}
\begin{verbatim}
EXECUTION TREE
==============
1 main [agent_b8845adf] (134.4s | 95752 tokens (cached 0.57%) | $0.185 | Success)
   [Stats] Events: 1 tool calls, 6 messages
   [Assistant] t+0.1s | "I'll orchestrate the solution to this problem foll..."
   1.3 reasoning_agent [agent_20fccbc1] (12.0s | 5673 tokens (cached 0.00%) | $0.013 | Success)
      [Stats] Events: 1 messages
      [Assistant] t+9.8s | "Below is an explanation that addresses the three p..."
   [Assistant] t+21.1s | "The reasoning agent has provided excellent analy..."
   1.2 solve_problem [agent_11804c1f] (17.6s | 54822 tokens (cached 0.68%) | $0.085 | Success)
      [Stats] Events: 3 tool calls, 4 messages
      [Assistant] t+26.1s | "Let me solve this step by step.1. UNDERSTANDING..."
      [Tool] calculate | 0.0s → Success
      [Assistant] t+32.4s | "Therefore there are floor(404.8) = 404 complete..."
      [Tool] return_result | 0.0s → Success
      [Assistant] t+42.1s | "<COMPLETE></COMPLETE>..."
   [Assistant] t+43.7s | "The problem solver has provided a solution..."
   1.1 software_developer [agent_0117a7ef] (69.7s | 208122 tokens (cached 0.68%) | $0.321 | Success)
      [Stats] Events: 8 tool calls, 10 messages
      [Assistant] t+49.3s | "I'll help solve this mathematical game problem thr..."
      [Tool] execute_command | 0.0s → Success
      [Assistant] t+54.6s | "<TOOL_CALL><TOOL_NAME>start_lsp_server</TOOL_NA..."
      [Tool] start_lsp_server | 2.2s → Success
      [Assistant] t+59.9s | "Now let's create our main implementation file:..."
      [Tool] overwrite_file | 0.0s → Success
      [Assistant] t+70.4s | "Now let's create a test script to verify our imp..."
      [Tool] overwrite_file | 0.0s → Success
      [Assistant] t+76.8s | "Now let's run our test script to verify everythi..."
      [Tool] execute_command | 0.0s → Success
      [Assistant] t+80.9s | "I notice that there's an issue with our pattern..."
      1.1.1 reasoning_agent [agent_8a69d943] (11.3s | 10231 tokens (cached 0.00%) | $0.017 | Success)
         [Stats] Events: 1 messages
         [Assistant] t+87.2s | "The problem comes from our backward induction func..."
      [Assistant] t+98.5s | "Based on the reasoning agent's analysis, let's u..."
      [Tool] overwrite_file | 0.0s → Success
      [Assistant] t+107.8s | "Now let's run our test script again to verify th..."
      [Tool] execute_command | 0.0s → Success
      [Assistant] t+112.0s | "Perfect! Our implementation now correctly:1. I..."
      [Tool] return_result | 0.0s → Success
      [Assistant] t+116.8s | "<COMPLETE></COMPLETE>..."
   [Assistant] t+119.0s | "Let's verify the software agent's answer since it di..."
   1.4 reasoning_agent [agent_6cd34374] (5.0s | 12025 tokens (cached 0.00%) | $0.016 | Success)
      [Stats] Events: 1 messages
      [Assistant] t+124.1s | "Answer synthesis: Our analysis starts..."
   [Assistant] t+129.1s | "Let's submit our final answer:<TOOL_CALL><TO..."
   [Tool] submit_answer | 0.0s → Success
   [Assistant] t+132.1s | "<COMPLETE></COMPLETE>..."

Total Duration: 249.9s
Total Tokens: 407766 (of which cached 232952)
Total Cost: $0.712
\end{verbatim}
\end{tracebox}
\end{adjustbox}
}

The above shows a representative example of an agent solving a problem. The problem at hand is a maths problem, and the execution begins with a reasoning agent (o3 mini) which devises a solution method. This is then passed to a `problem solving' agent which attempts to solve the problem. Then, the main agent invokes a `software developer' agent to write a script to numerically verify the solution. The software developer agent gets stuck after attempting to run the test script, and invokes the reasoning agent to help debug the problem. After applying the fix, the test script runs correctly and yields a solution that differs from the first problem solver's answer. To consolidate the two, the main agent invokes another reasoning agent, which sides with the empirically derived solution, which is finally submitted as the answer by the main agent.

{
\begin{adjustbox}{max width=\linewidth, center}
\begin{tracebox}
\begin{verbatim}
EXECUTION TREE
==============
1 main [agent_5e260c4e] (617.5s | 106497 tokens (cached 0.30%) | $0.299 | Success)
   [Stats] Events: 1 tool calls, 5 messages
   [Assistant] t+0.1s | "I'll help orchestrate the solution to this problem..."
   1.1 archive_explorer [agent_534a61ec] (88.6s | 372576 tokens (cached 0.16%) | $1.205 | Success)
      [Stats] Events: 7 tool calls, 10 messages
      [Assistant] t+5.6s | "I'll follow the procedure step by step to analyze..."
      [Tool] compare_agent_iterations | 0.0s → Success
      [Assistant] t+9.7s | "Let's also look at some of the best and worst pe..."
      [Tool] best_problems | 0.0s → Success
      [Assistant] t+13.2s | "Let me also look at some of the worst performing..."
      [Tool] worst_problems | 0.0s → Success
      [Assistant] t+16.7s | "Let me inspect the project documentation..."
      [Tool] open_file | 0.0s → Success
      [Assistant] t+22.1s | "<TOOL_CALL><TOOL_NAME>open_file</TOOL_NAME><TO..."
      [Tool] open_file | 0.0s → Success
      [Assistant] t+27.4s| "Let me also look at the current iteration's desc..."
      [Tool] open_file | 0.0s → Success
      [Assistant] t+33.6s| "Having reviewed the documentation and cu..."
      1.1.2 reasoning_agent [agent_d4553475] (14.8s | 81577 tokens (cached 0.00%) | $0.097 | Success)
         [Stats] Events: 1 messages
         [Assistant] t+44.0s| "I updated description.txt and agent_change_log.md..."
      [Assistant] t+58.1s | "Let me examine this suggestion from the reasoni..."
      1.1.1 reasoning_agent [agent_ccad2709] (5.4s | 81442 tokens (cached 0.02%) | $0.090 | Success)
         [Stats] Events: 1 messages
         [Assistant] t+72.0s | "I’ve updated both description.txt and agent_change..."
      [Assistant] t+77.4s | "Based on the analysis from both reasoning agents..."
      [Tool] return_result | 0.0s → Success
      [Assistant] t+88.7s | "<COMPLETE></COMPLETE>..."
   [Assistant] t+94.1s | "The archive explorer has made a specific suggest..."
   1.2 software_developer [agent_6345b78d] (331.2s | 1388672 tokens (cached 0.22%) | $4.341 | Success)
      [Stats] Events: 21 tool calls, 22 messages
      [Assistant] t+109.7s | "Let's first look at key documentation files:<T..."
      [Tool] open_file | 0.0s → Success
      [Assistant] t+114.4s | "Now let's look at any existing diff-related code..."
      [Tool] open_file | 0.0s → Success
      [Assistant] t+118.3s | "Let's also look at existing utils and __init__ f..."
      [Tool] open_file | 0.0s → Success
      [Assistant] t+123.1s | "Now that I understand the context better, let me..."
      [Tool] execute_command | 0.0s → Success
      [Assistant] t+274.6s | "Now let's run the tests to verify the implementa..."
      [Tool] execute_command | 0.5s → Success
      [Assistant] t+281.9s | "Oops, looks like I need to fix the test case. Th..."

      ... truncated ...

      [Tool] diff_patch_edit | 0.0s → Success
      [Assistant] t+356.8s | "Now let me run the tests again:<TOOL_CALL><T..."
      [Tool] execute_command | 0.5s → Success
      [Assistant] t+365.3s | "Great! Now that the implementation and tests ar..."
      [Tool] diff_patch_edit | 0.0s → Success
      [Assistant] t+375.6s | "Now let me update description.txt:<TOOL_CALL>..."
      [Tool] open_file | 0.0s → Success
      [Assistant] t+382.6s | "<TOOL_CALL><TOOL_NAME>diff_patch_edit</TOOL_NAM..."
      [Tool] diff_patch_edit | 0.0s → Success
      [Assistant] t+392.4s | "Finally, let me update agent_change_log.md:<..."
      [Tool] diff_patch_edit | 3.6s → Success
      [Assistant] t+415.4s | "Let me verify that I've completed all the requ..."
      [Tool] return_result | 0.0s → Success
      [Assistant] t+428.0s | "<COMPLETE></COMPLETE>..."
   [Assistant] t+434.6s | "The software developer has successfully impleme..."
   1.3 software_developer [agent_ae600007] (167.0s | 194915 tokens (cached 0.54%) | $0.390 | Success)
      [Stats] Events: 9 tool calls, 10 messages
      [Assistant] t+442.9s | "I'll help test the DiffSummarizer implementation b..."
      [Tool] open_file | 0.0s → Success
      [Assistant] t+447.8s | "Now let's create a directory for our test and se..."
      [Tool] execute_command | 0.0s → Success
      [Assistant] t+456.7s | "Now let's execute the agent command to run our t..."
      [Tool] execute_command | 0.0s → Success
      [Assistant] t+460.5s | "I apologize for the error. We need to properly s..."
      [Tool] execute_command | 120.3s → Failed
      [Assistant] t+585.8s | "I see that the command is taking longer than exp..."
      [Tool] open_file | 0.0s → Success
      [Assistant] t+588.9s | "Let's try a direct test of the DiffSummarizer. W..."
      [Tool] execute_command | 0.0s → Success
      [Assistant] t+595.5s | "Now let's run our test script:<TOOL_CALL><TOO..."
      [Tool] execute_command | 0.5s → Success
      [Assistant] t+598.5s | "Perfect! Let's analyze what we've found:1. The..."
      [Tool] return_result | 0.0s → Success
      [Assistant] t+608.3s | "<COMPLETE></COMPLETE>..."
   [Assistant] t+609.8s | "Great! The implementation and testing of the Di..."
   [Tool] submit_answer | 0.0s → Success
   [Assistant] t+615.2s | "<COMPLETE></COMPLETE>..."
\end{verbatim}
\end{tracebox}
\end{adjustbox}
}

The above shows a characteristic meta-improvement run: the agent starts by invoking the `archive explorer' sub-agent, which has access to a number of tools to facilitate archive analysis. Note that this is using exactly the same simple, generic and extensible sub-agent class as the other base agents in the system. The archive explorer compares different iterations, looks at the best performing benchmark problems in a given iterations to understand success cases, as well as the worst performing benchmark problems at a given iteration to understand the failure modes. It then views some files to understand the agetn's implementation and documentation, and then invokes two reasoning agents to help it reason over this information. The result of this agent call is a recommendation to the main agent about what to work on.

The main agent then invokes the `software developer' sub-agent with the
instruction to implement the feature. The coding sub-agent first looks at
relevant files and documentation, implement the feature, and then enters into a
debugging loop as it fixes the bugs in its implementation. About 3 minutes
later, it successfully gets all the tests (that it wrote) to pass, documents the
change in the agent change log, and returns to the main agent.

The main agent then invokes a second software developer sub-agent to
independently verify the implementation of the feature, which passes after an
initial issue, and the agent returns a description of the newly functioning tool
to the main agent which then exits.

\section{Function Calling Interface}%
\label{sec:function_calling}

The \acr{LLM} function calling interface allows the \acr{LLM} to invoke tools and sub-agents at run-time without pre-defined control flow.
There are many approaches to implementing this, from constrained generation, to parsing the stream of agent token generations without interruption, to simply relying on \acr{LLM} providers' or inference frameworks' native function or tall calling implementations.

For maximum flexibility, we adopt an un-constrained structured generation
approach, where the \acr{LLM} outputs \acr{XML}-formatted generations, where the
closing tag is registered as a stop token. Inspecting the stop reason, we can
then detect whether a tool call or an agent call occurred, parse the content,
and run the tool implementation.

For instance, here is an example tool calling syntax:
\begin{verbatim}
<TOOL_CALL>
<TOOL_NAME>tool_name</TOOL_NAME>
<TOOL_ARGS>
<arg1>value1</arg1>
<arg2>value2</arg2>
</TOOL_ARGS>
</TOOL_CALL>
\end{verbatim}
where \texttt{</TOOL\_CALL>} is registered as a stop token. The
\texttt{tool\_name} is then parsed first and is used to index into a set of tools
made available to the calling agent, following which the tool arguments are
parsed into a dict, and the tool class is instantiated.

We choose \acr{XML} over \acr{JSON} as the tool calling format since it does not
require escaping string literals, which can become burdensome when including
file content verbatim as an argument.

\section{Additional Result Details}%
\label{subsec:result_details}

Here is the sequence of updates the agent made, corresponding to the plot in Figure~\ref{fig:performance}.

\begin{enumerate}
    \item \textbf{Smart Editor}: Implemented a new SmartEditor tool to
intelligently select and execute the optimal file editing strategy based on edit
characteristics.
    \item \textbf{Quick Overwrite Tool}: to address performance issues with file editing
        operations to improve on full file overwrites; reducing token usage.
    \item \textbf{Diff-Enhanced Smart Editor}: adds intelligent diff-based
        strategy selection and improved pattern-based editing.
    \item \textbf{Simplified DiffVerifier Tests}: improved the developer
        experience to improve maintainability
    \item \textbf{Code Context Summarizer}: Added a new CodeContextSummarizer
        tool that efficiently extracts and summarizes code context using
        ripgrep. Intended to make codebase exploration more efficient.
    \item \textbf{SmartEditor Input Normalizer}: Added an InputNormalizer
        component to improve edit operation efficiency and reliability.
    \item \textbf{Enhanced File Edit Verification}: Implemented a comprehensive
        file state tracking system to improve edit operation efficiency.
    \item \textbf{Minimal Diff Output Optimization}: Implemented optimizations
        to reduce token usage in file editing operations by minimizing diff
        context output.
    \item \textbf{AST Symbol Locator}: Integrated an AST-based symbol locator
        tool for precise code navigation.
    \item \textbf{Optimized File Edit Routing}: Implemented explicit task type
detection and routing optimization in MainAgent to improve handling of file
editing tasks.
    \item \textbf{Enhanced File Size Optimization}: Added file size thresholding
        of 100KB to optimize handling of large files and reduce processing
        overhead on big changes.
    \item \textbf{Hybrid Symbol Locator}: Implemented a hybrid symbol location
        approach that combines ripgrep-based candidate identification with AST
        verification.
    \item \textbf{Context-Sensitive Diff Minimization}: Completed the
        integration of context-sensitive diff minimization into the SmartEditor
        tool. Uses AST parsing for Python code changes and pattern-based
        analysis for other file types. Intends to reduce token usage by
        minimizing change context.
    \item \textbf{Optimized AST Parsing in Hybrid Symbol Locator}: Implemented
        targeted AST parsing in hybrid\_symbol\_locator.py to process only
        relevant code regions instead of whole files.
\end{enumerate}